\documentclass[sigconf]{acmart}

\usepackage{booktabs} %
\usepackage{subfig}
\usepackage{color}
\usepackage{multirow}
\usepackage{balance}
\usepackage{booktabs}

\AtBeginDocument{%
  \providecommand\BibTeX{{%
    \normalfont B\kern-0.5em{\scshape i\kern-0.25em b}\kern-0.8em\TeX}}}

\newcommand{\firstwwj}{\textcolor{black}}%
\newcommand{\wwj}{\textcolor{black}}

\newcommand{\ys}{\textcolor{black}}%

\copyrightyear{2020} 
\acmYear{2020} 
\setcopyright{acmcopyright}\acmConference[MM '20]{Proceedings of the 28th ACM International Conference on Multimedia}{October 12--16, 2020}{Seattle, WA, USA}
\acmBooktitle{Proceedings of the 28th ACM International Conference on Multimedia (MM '20), October 12--16, 2020, Seattle, WA, USA}
\acmPrice{15.00}
\acmDOI{10.1145/3394171.3413953}
\acmISBN{978-1-4503-7988-5/20/10}

\begin{document}
\fancyhead{}

\title{\wwj{From Design Draft to Real Attire: Unaligned Fashion Image Translation}}

\author{Yu Han}
\affiliation{%
  \institution{Wangxuan Institute of Computer Technology, Peking University}
}
\email{vickyhan@pku.edu.cn}

\author{Shuai Yang}
\affiliation{%
  \institution{Wangxuan Institute of Computer Technology, Peking University}
}
\email{williamyang@pku.edu.cn}

\author{Wenjing Wang}
\affiliation{%
  \institution{Wangxuan Institute of Computer Technology, Peking University}
}
\email{daooshee@pku.edu.cn}

\author{Jiaying Liu}
\authornotemark[0]
\authornote{Corresponding Author. This work was supported by National Natural Science Foundation of China under contract No.61772043, and Beijing Natural Science Foundation under contract No. 4192025 and No.L182002.}
\affiliation{%
  \institution{Wangxuan Institute of Computer Technology, Peking University}
}
\email{liujiaying@pku.edu.cn}

\begin{abstract}
  Fashion manipulation has attracted growing interest due to its great application value, which inspires many researches towards fashion images. However, little attention has been paid to fashion design draft. In this paper, we study a new unaligned translation problem between design drafts and real fashion items, whose main challenge lies in the huge misalignment between the two modalities. We first collect paired design drafts and real fashion item images without pixel-wise alignment. To solve the misalignment problem, our main idea is to train a sampling network to adaptively adjust the input to an intermediate state with structure alignment to the output. Moreover, built upon the sampling network, we present design draft to real fashion item translation network (D2RNet), where two separate translation streams that focus on texture and shape, respectively, are combined tactfully to get both benefits. D2RNet is able to generate realistic garments with both texture and shape consistency to their design drafts. We show that this idea can be effectively applied to the reverse translation problem and present R2DNet accordingly. Extensive experiments on unaligned fashion design translation demonstrate the superiority of our method over state-of-the-art methods. Our project website is available at: {\url{https://victoriahy.github.io/MM2020/}}.
\end{abstract}



\maketitle

\section{Introduction}

\begin{figure}[t]
  \centering
  \includegraphics[width=0.96\linewidth]{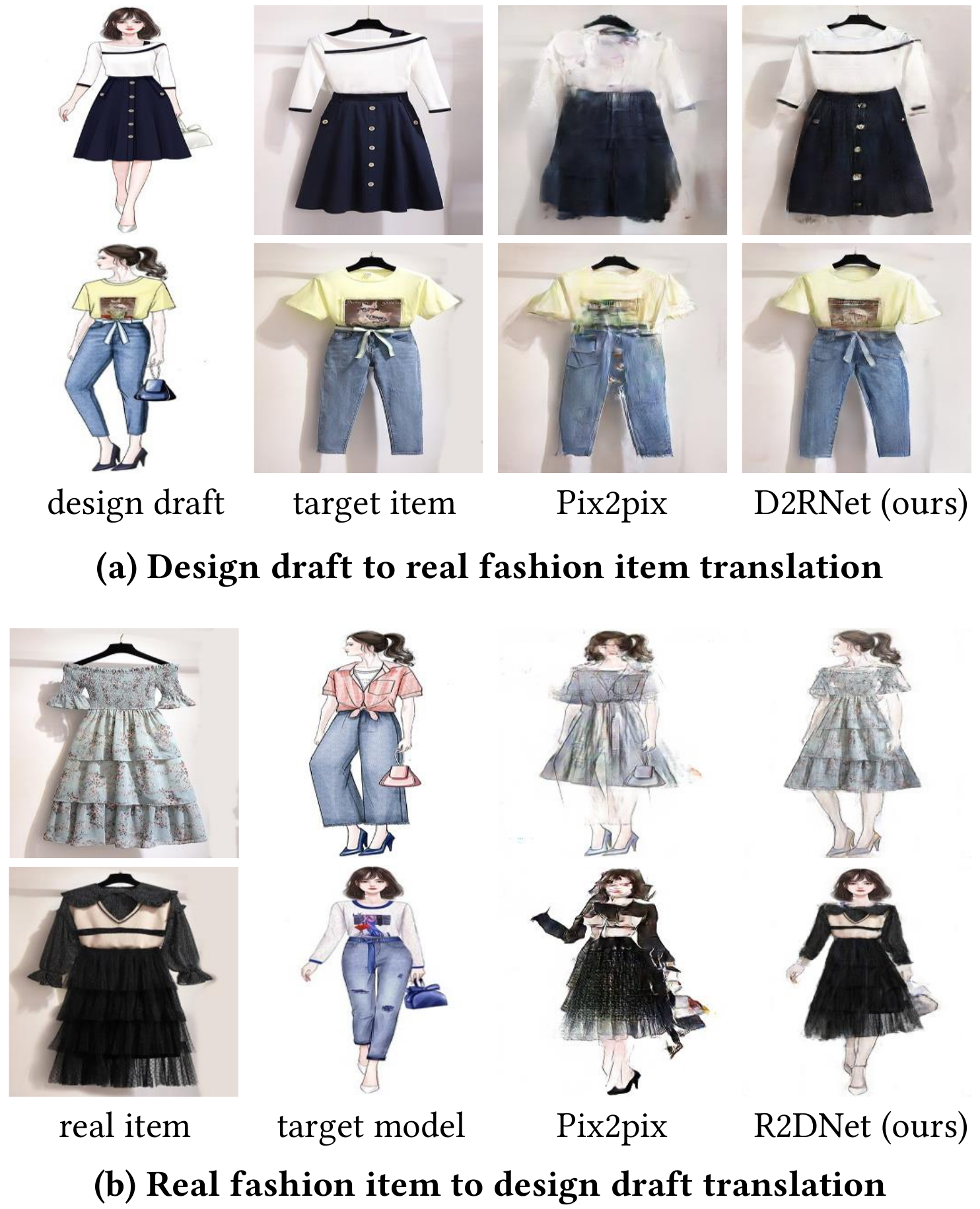}
  \caption{Our method allows translations between unaligned design drafts and real fashion items.  Compared with Pix2pix~\cite{isola2017image}, our method generates accurate shapes and preserves vivid texture details.}
  \label{fig:teaser1}
\end{figure}

\ys{
The outbreak of people's pursuit of personalized expression \wwj{has} spawned new demands for customization of products, especially for the fashion item design such as the clothing, hats and accessories.
Thanks to the recent development of generative adversarial networks (GANs), the intelligent image-based fashion editing becomes possible.
The model can automatically generate the corresponding clothing according to the user's language guidance~\cite{AMGAN2019,zhu2017be}, or recommend the clothing matching according to the user's favorite~\cite{fashiondesign}, or even provide virtual try-on service~\cite{han2017viton,hsiao2019fashionplus,wang2018toward}, \textit{etc}.
These models enable normal users to convert their creative ideas into corresponding fashion items, which greatly reduces the difficulty for user customization, and is of certain entertainment value and huge commercial value.}

\begin{figure}[t]
  \centering
  \includegraphics[width=0.98\linewidth]{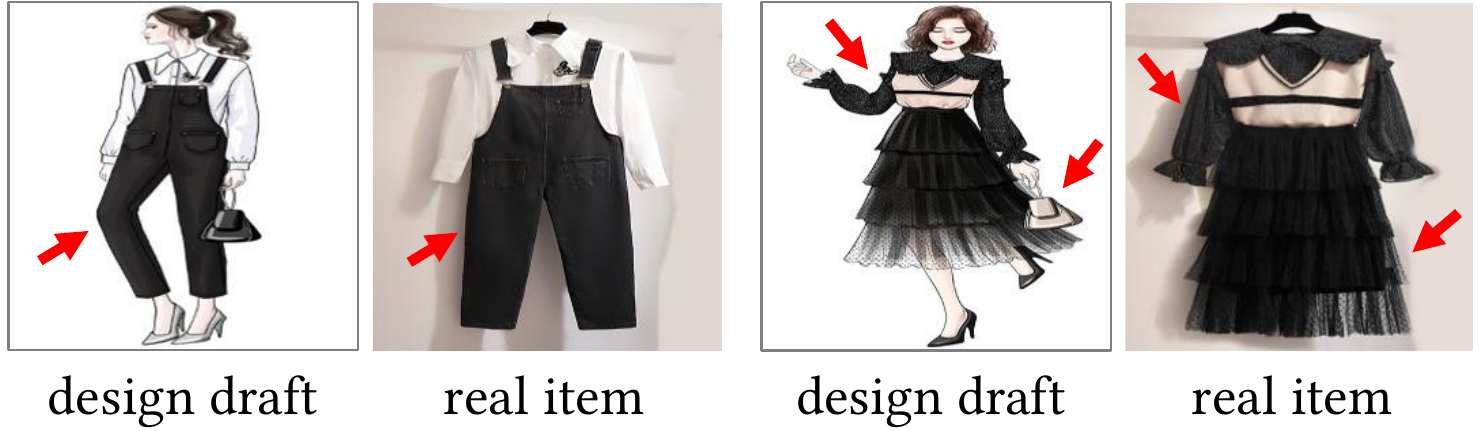}
  \caption{Two examples in our collected dataset. There is \wwj{an} evident structural discrepancy between the design drafts and real fashion items due to the exaggerated proportion of the human body and the pose of the model in design drafts as indicated by the red arrows.}
  \label{fig:teaser}
\end{figure}

\ys{For fashion design, designers tend to begin their \wwj{ideas} with design drafts. The design draft can accurately reflect the appearance of \wwj{a} fashion item, and more importantly, can be easily drawn and modified. Thus fashion design drafts are crucial for fashion analysis and editing.
However, to the best of our knowledge, most existing studies focus on highly abstract labels or languages. Much less has been done to the design drafts.
}
Furthermore, there is no related design draft dataset so far to facilitate the corresponding researches.
\ys{This practical requirement motivates our work in this paper: we investigate a new design draft to real fashion item translation problem, which would facilitate the preview of the fashion designs for designers.}

\ys{The goal of design draft to real fashion item translation is} to obtain the fashion item images which have realistic styles in both shape and texture to original design drafts as shown in Fig.~\ref{fig:teaser1}(a). When users modify the design draft, they can simply view the corresponding manipulations on real fashion items.
\ys{To fill this research blank}, we first collected a dataset containing paired design drafts and real fashion items. The challenges of our problem lie in two aspects.
First, as shown in Fig.~\ref{fig:teaser}, there is an evident structural discrepancy between the design draft and its real counterpart. For example, the models in both examples have exaggerative legs-to-upper-body ratio \wwj{and complex pose} which cause deformation to the clothes. Therefore, the unaligned structures make it difficult to synthesize real fashion items with the details its design drafts represent. 
Second, in contrast to rigid objects, clothing items contain rich textures, which are severely distorted due to the poses of the models, making it difficult to simultaneously synthesize both item shapes and textures without introducing unwanted artifacts.

\ys{These two challenges greatly limit the performance of existing image-to-image translation models.}
Previous methods such as Pix2pix \cite{isola2017image}, Pix2pix-HD \cite{wang2018pix2pixHD}, Art2Real \cite{tomei2019art2real}, and SPADE \cite{park2019SPADE} require their input and output to be strictly aligned in pixel level. As shown in Fig.~\ref{fig:teaser1} and Fig.~\ref{fig:com1}, they are not adaptive to our task, and it is hard for them to learn the structure mapping between design drafts and real fashion items.
Many fashion-related translation as well as the guided image-to-image translation~\cite{wu2019relgan} researches take use of segmentation maps, parsing labels, or human pose landmarks to improve the performance.
However, since our real fashion item images do not contain models, it brings considerable difficulties for networks to locate landmarks or class labels.
In addition, the design draft is smooth and has less \wwj{texture} compared to its real counterpart, which leads to inaccurate segmentation results.
In fact, we have tried the powerful parsing methods~\cite{gong2017look,liang2018look} as well as the landmark detection method \cite{liu2016fashionlandmark} and observed poor results on our dataset. \ys{It means \wwj{that} it is not realistic to obtain the aforementioned guidance information to improve the translation performance as other fashion-related researches do.}

To address this issue, we develop structure-aware D2RNet to solve the design draft to real fashion item translation problem. In particular, the proposed approach explores two subnetworks, \textit{i.e.}, detail preservation network and shape correction network\wwj{,} to generate detail and shape for fashion items, respectively. Both \wwj{leverage} a sampling layer to adaptively warp the input design drafts to match the real items. We show that by simply conditioning the discriminator with design drafts before or after warping, the two generators can effectively learn to focus on the texture details and the overall shapes, respectively.
To combine the two merits, we utilize a fusion network to fuse the results from two streams to better reconstruct shape, texture and details. Through our network, designers can manipulate on fashion item's design draft form and get the photorealistic form immediately. 
In addition, our model can be easily adapted to an exemplar-guided framework for real fashion item to design draft translation, \textit{i.e.}, reversed D2RNet, as shown in Fig.~\ref{fig:teaser1}(b), which can be used for \wwj{the} display of e-shops.
\wwj{Experimental results} demonstrates the superiority of our method in both detail preservation and shape adjustments over state-of-the-art image-to-image translation methods.
In summary, the contribution of our work is as follows:

\begin{itemize}
\item We raise a new problem of unaligned design draft to real fashion item translation. To the best of our knowledge, we are the first to address this new problem \wwj{of both fashion item shape preservation and realistic texture detail generation.}
\item We reduce the structure gap dramatically through introducing a saliency-based sampling layer with \wwj{a} standard image-to-image translation model.
\item \ys{We propose a two-stream network that works for both design draft to real fashion item translation and the reverse translation, where shapes and textures are transferred specifically in each stream and both advantages are combined through stream fusion. }
\end{itemize}

\begin{figure*}[t]
  \centering
  \includegraphics[width=0.9\linewidth]{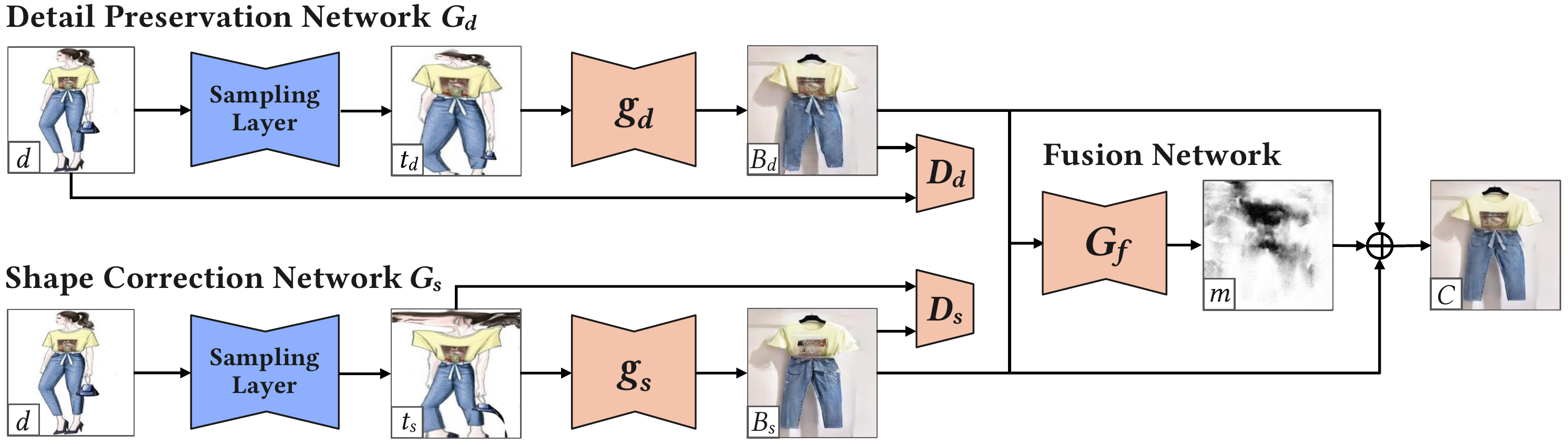}
  \caption{Our D2RNet framework. \firstwwj{First, the detail preservation network $G_d$ and the shape correction network $G_s$ translate texture and shape, respectively. Then, the two streams are combined by the fusion network to generate the final output. Notice that the result of $G_d$ has more detailed patterns of the T-shirt but an irregular shape of the pants, while the shape of the pants in $G_s$ is appropriate. The fused result $C$ has both detailed patterns and fine shape.}}
  \label{fig:framework1}
\end{figure*}

\section{Related Work}
\label{sec:related_work}

\subsection{Image-to-Image Translation}
Many computer vision \firstwwj{tasks} can be viewed as learning a mapping between images from one domain to another, such as super resolution, image denoising, image style transfer and image colorization. Isola \textit{et al.} \cite{isola2017image} first summarized these tasks as image-to-image translation, and proposed a universal framework named Pix2pix.
Since then, a lot of works have been done to improve Pix2pix in terms of unpaired training such as CycleGAN \cite{CycleGAN2017}, multi-model such as MUNIT \cite{huang2018munit} and multi-domain such as StarGAN \cite{choi2018stargan}, \textit{etc}.

Guidance can provide not only extra information to improve the performance but also a user editing interface for higher flexibility, which is an important topic for image-to-image translation.
Recently, a variety of guidance has been designed, such as class label, segmentation map, image example, and attention, \textit{etc}.
For class label guided tasks, Wu \textit{et al.} \cite{wu2019relgan} introduced RelGAN for multi-domain image-to-image translation using relative attributes. Zhang \textit{et al.} \cite{jichao2018} proposed a novel approach SG-GAN that can perform image translation in the sparsely grouped datasets.
For segmentation guided tasks, Wang \textit{et al.} \cite{wang2018pix2pixHD} gained a high-resolution result through a novel adversarial loss as well as multi-scale generator and discriminator architectures. Tomei \textit{et al.} \cite{tomei2019art2real} proposed a method to translate artworks to real images based on constructions of efficient memory banks. Park \textit{et al.} \cite{park2019SPADE} extended spatially-adaptive denormalization for photorealistic tasks.
For texture patch guided tasks, TextureGAN \cite{xian2017texturegan} first examined the texture control through a local texture loss. AlBahar \textit{et al.} \cite{albahar2019guided} then presented a bi-directional feature transformation (bFT) scheme for this task. For sketch guided tasks, Yang \textit{et al.} \cite{Yang2020Deep} provided face editing based on the structural information of sketches.
For exemplar-guided tasks, Pix2pixSC \cite{pix2pixSC2019} exploited style-consistency discriminator and an adaptive semantic consistency loss
\firstwwj{to maintain style consistency with the exemplar.}

The above methods require the input to be strictly aligned with the output in pixel wise, or else the result will be not ideal.
\ys{However, in many cases, well-aligned data are hard to obtain, especially for design drafts and real fashion items.
In this paper, we introduce a structure-aware sampling layer to adaptively establish the structure mappings between the two modalities to solve this problem.}

\subsection{Fashion Synthesis}

In the field of fashion synthesis researches, fashion represents the styles of clothing and appearance.
Recently, there has been a growing interest in fashion synthesis, such as fashion design, fashion image manipulation, virtual try-on and so on.
FashionGAN \cite{zhu2017be} and AMGAN \cite{AMGAN2019} are introduced to conduct text-guided fashion item manipulation such as long sleeves to short sleeves while preserving other parts of the clothes.
Fashion++ \cite{hsiao2019fashionplus}, VITON \cite{han2017viton}, CP-VITON \cite{wang2018toward} and E2E \cite{Song2010Pose,Song2020} can make adjustments to clothing outfit and realize the function of trying on clothes with one photo.

Different from existing researches, our work focuses on a new task of translating \firstwwj{between} design drafts \firstwwj{and} fashion items. To resolve this problem, \ys{we collected a paired but unaligned dataset and designed a novel two-stream image translation framework.}

\section{D2RNet: Design Draft to Real Fashion Item Translation}

The research focus of mapping design draft to real fashion item lies in two aspects: generating detailed textures and adjusting shapes.
Let \firstwwj{$d$} be an input image from a design draft domain, we have its corresponding ground truth photo \firstwwj{$r$}. Our particular challenge is that $d$ and $r$ \firstwwj{are} not pixel-wisely aligned. The structure of clothes in $d$ is exaggerated out of portion.
Meanwhile, the model has various poses, causing severe deformations to the clothes.
Even if we remove the body part and enlarge the cloth part to equal proportions, it does not match the real clothing items in a normal state. To make it worse, existing methods \cite{gong2017look,liang2018look,liu2016fashionlandmark} unfortunately can not provide accurate landmark or semantic segmentation of the design drafts and the real fashion items as guidance.

\ys{
To solve these problems, we present a novel D2RNet, containing two structure-aware alignment networks and a fusion network $G_f$.  An overview of our model is presented in Fig.~\ref{fig:framework1}. The two structure-aware alignment networks, \textit{i.e.}, the detail preservation network $G_d$ and \firstwwj{the} shape correction network $G_s$\firstwwj{,} are adept with the texture translation and shape translation, respectively. Both \firstwwj{are} equipped with a saliency-based sampling layer $S$ to tactfully avoid the lack of guidance. The fusion network \firstwwj{fuses} results from two streams to combine the two merits.
}
In the following, we represent each \firstwwj{component} of our networks in detail.

\subsection{Detail Preservation Network}
\label{sec:Gr}

To make the details of $d$ in an alignment to $r$, our detail preservation network $G_d=g_d\circ S$ contains a saliency-based sampling layer $S$ \cite{recasens2018learning} and a generator $g_d$.
\ys{
$g_d$ follows U-Net as in Pix2pix. Meanwhile, the saliency-based sampling layer aims to predict a task-relevant saliency map, which is then used to guide the sampling of the pixels from the input to achieve a saliency-based image distortion. This differentiable layer can be trained altogether with $g_d$ in an end-to-end fashion so that it learns to efficiently estimate how to sample from $d$ in order to align the ideal output. Please refer to~\cite{recasens2018learning} for implementation details of $S$. The advantage of the saliency-based sampling layer is that it can roughly align the data without losing any information from the input or introducing any network-generated artifacts.
}

A discriminator $D_d$ is added to \firstwwj{judge} the authenticity of the generated image and whether it matches the draft. It is important to note that \textit{the conditional input to $D_d$ is original $d$ because $d$ has all design draft details with no distortion, which help $G_d$ focus on the texture details.} We apply the $L1$ loss, perceptual loss $L_{per}$ \cite{johnson2016perceptual} to the ground truth $r$ and HingeGAN \cite{miyato2018spectral} loss $L_{hinge}$:

\begin{equation}
\begin{aligned}
L_{D_d}=&~\mathbb{E}_{d,r}[max(0,1-D_d(d,r))]\\
+&~\mathbb{E}_{d}[max(0,1+D_d(d,G_d(d)))],
\end{aligned}
\end{equation}
\begin{equation}
L_{G_d}=-\mathbb{E}_d[D_d(d,G_d(d))].
\end{equation}

Based on our networks, the generation of $B_d=G_d(d)$ has not only design details presented in $d$, but also appropriate structure proportion. The final loss is formulated as:
\begin{equation}
L_d=L_{G_d}+L_{D_d}+\lambda_1 L_1+\lambda_2 L_{per}.
\end{equation}

\label{sec:text_style_transfer}
\begin{figure*}[t]
  \centering
  \includegraphics[width=0.87\linewidth]{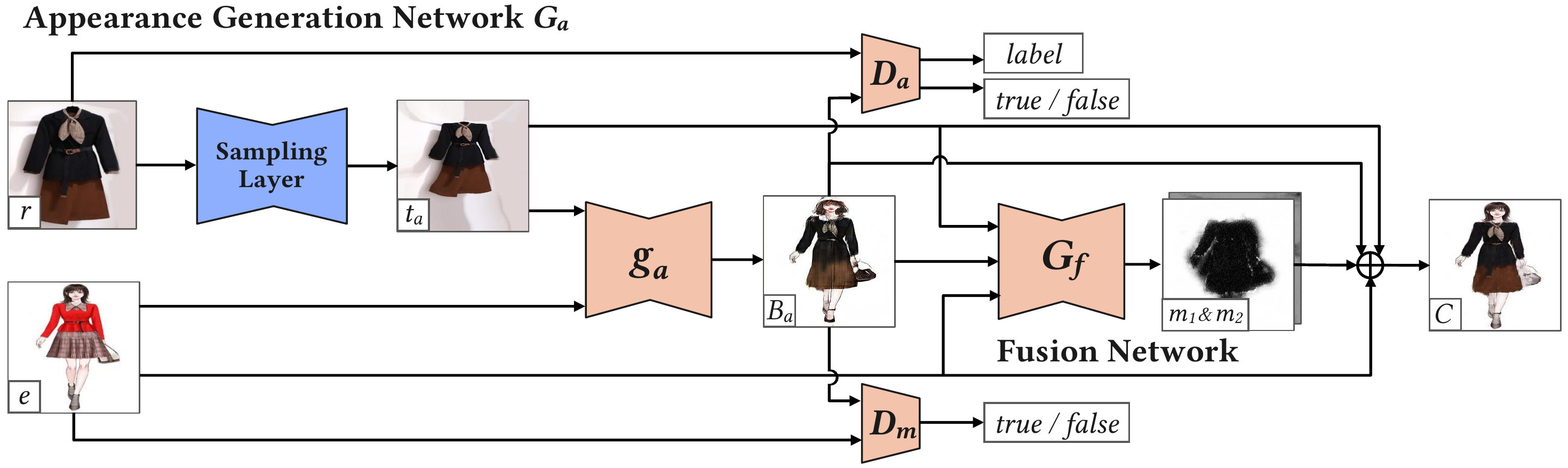}
  \caption{Our R2DNet framework. \firstwwj{The appearance generation network $G_a$ consists of a saliency-based sampling layer and a generator. Using the input real fashion item $r$ and the exemplary target model $e$, $G_a$ generates a preliminary design draft image $B_a$. This result is then refined with the help of $e$ and $t_a$ through the fusion network.}}
  \label{fig:framework2}
\end{figure*}

\subsection{Shape Correction Network}
\label{sec:attention}

We can see from the distortion image $t_d$ and the result $B_d$ that the detail preservation network mainly focuses on retaining design patterns and decorations of the drafts. This is because the sampling grid enlarges the center of clothes in the design draft.
Because U-Net transfers too much information about $t_d$ to the output $B_d$, those crooked margins in $t_d$ appear in $B_d$. To avoid this unwanted \firstwwj{artifact, we need to fix the shape distortion problem.}

\ys{One possible way is to add a subsequent Pix2pix model after $G_d$ to revise the shape, therefore forming a coarse-to-fine process as in many image translation models~\cite{wang2018pix2pixHD}.
However, we found by \firstwwj{experiments} that as the whole framework goes deeper, the details in the original draft are inevitably and severely lost. Even applying residual learning does not relieve the problem due to the great structure discrepancy. To obtain more information from the original input instead of the image we generate, we introduce a two-stream framework, where another stream of shape correction network is proposed.}

\ys{$G_s$ shares the same network architecture as $G_d$. The only difference is the conditional input to the discriminator. \textbf{Our key observation is that through changing the conditional image, the same network will pay attention to different aspects of the translation task.}}
\firstwwj{Interestingly we find that \textit{if we use the distortion image as the conditional input for $D_s$, the grid sampler will tend to adjust the shape of $d$} rather than preserving texture details.}
A possible explanation for it is that $t_s$ itself is \firstwwj{an} intermediate result of the generator, using it as the conditional input will drive itself as well as the final output to approach $r$ more broadly. However, when the conditional input is original design drafts, the discriminator will be more \firstwwj{focused} on the central pattern and less sensitive to scaling changes.
As for the loss function, the HingeGAN loss changes to
\begin{equation}
\begin{aligned}
L_{D_s}=&~\mathbb{E}_{d,r}[max(0,1-D_s(t_s,r))]\\
+&~\mathbb{E}_{d}[max(0,1+D_r(t_s,G_s(d)))],
\end{aligned}
\end{equation}
where $t_s=S(d)$ is the intermediate distortion result. And the final loss is formulated as:
\begin{equation}
L_s=L_{G_s}+L_{D_s}+\lambda_1 L_1+\lambda_2 L_{per}.
\end{equation}

\subsection{Fusion Network}
\label{sec:datacollect}

Since we now have two generated images, $B_d=G_d(d)$ with detail preservation to $d$ and $B_s=G_s(d)$ with clear edges and shapes, we would like to combine them with both advantages and fix each other's artifacts. We estimate a mask $m$ through putting both $B_d$ and $B_s$ into a U-Net based mask generator $G_f$. The final result of our networks is:
\begin{equation}
C=B_s\otimes m+ B_d\otimes(1-m),
\end{equation}
where $\otimes$ is the element-wise multiplication operation.

However, the details of $B_d$ and edges of $B_s$ are not so perfectly aligned to the ground truth $r$. If we simply apply $L1$ loss to the mask's generation, the combination will be \firstwwj{blurred}.
To handle this \firstwwj{misalignment}, we use contextual loss $L_{cx}$ \cite{mechrez2018Learning,mechrez2018contextual} to maintain both the detail of $B_d$ and the clear edge of $B_s$.
Let $l$ \firstwwj{denote} the layer of the perceptual network VGG19 \cite{vgg}:
\begin{equation}
L_{c}=L_{cx}(C,r,l_t)+L_{cx}(C,r,l_s),
\end{equation}
where $l_t=conv3\_2$ to capture detail and $l_s=conv4\_2$ to capture general shape. Meanwhile, the generator is supposed to learn a coarse-grained representation of the mask.
To make the mask more regular, we adopt Total Variation~(TV) loss as:
\begin{equation}
L_{tv}=\sum_{i,j}((m_{i,j-1}-m_{i,j})^2+(m_{i,j+1}-m_{i,j})^2),
\end{equation}
where $m_{i,j}$ denotes the pixel at coordinate $(i,j)$. The final objective function of our fusion network is:
\begin{equation}
L_{f}=L_c+\lambda_3 L_{tv}.
\end{equation}

\section{R2DNet: Real Fashion Item to Design Draft Translation}
\label{sec:experiment}
The real fashion item to design draft translation is examplar-based.
Since design drafts in our dataset have additional model and pose information, this information should be fed into the inverse D2RNet, \textit{i.e.}, R2DNet for guidance.
As a result, the real fashion item to design draft translation becomes a problem of exchanging the clothing item on the model.
\ys{
To define our problem more clearly, we select seven classes from our dataset based on the model, with each class containing design drafts of similar model and pose and their corresponding real fashion item images.
Let $R_i$ and $D_i$ denote the real fashion item subdomain and design draft subdomain corresponding to the $i$-th class, respectively.
For the $i$-th class, let \firstwwj{$r$} be an input image from $R_i$, we have its corresponding ground truth $d$. \firstwwj{A} guidance image $e$ for training is randomly picked in $D_i$. And our goal is to replace the clothing item of $e$ with \firstwwj{that of} $r$ to approach the target $d$.
However, we are still faced with two challenges. First, the model in the same class is not well \firstwwj{pixel-wisely aligned}. Second, clothing exchange requires the network to create information that is not provided by $e$. For example, if we switch a model's pants to skirt, our network needs to generate model's leg which is covered by the pants in $e$.
We show that by adding high-level label guidance, our two-stream framework can be effectively applied to this challenging task.
}
\firstwwj{Our R2DNet is similar to the D2RNet except that one stream is directly replaced with the guidance image.}
\ys{An overview of our network is presented in Fig.~\ref{fig:framework2}. Our R2DNet generates a preliminary result from \firstwwj{an} exemplar-based structure-aware network and combines this result with the exemplar and the distorted $r$ to obtain the final result $C$. In the following, we represent each \firstwwj{component} of our networks in detail.}

\begin{figure*}[t]
  \centering
  \includegraphics[width=0.87\linewidth]{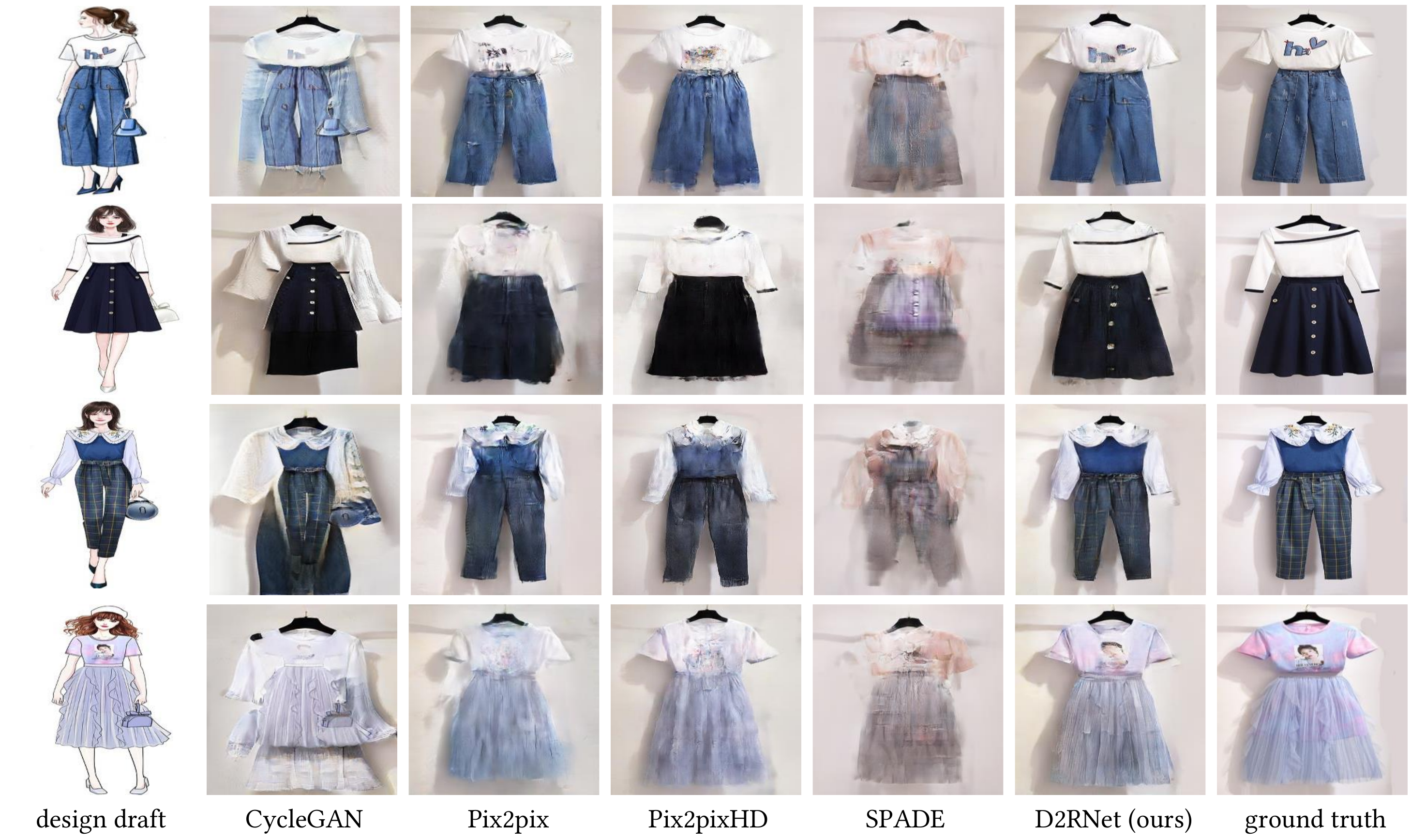}
  \caption{Our D2RNet compared with CycleGAN \cite{CycleGAN2017}, Pix2pix \cite{isola2017image}, Pix2pixHD \cite{wang2018pix2pixHD} and SPADE \cite{park2019SPADE}.}
  \label{fig:com1}
\end{figure*}

\subsection{Appearance Generation Network}

\ys{Considering the deformation caused by the model pose, we need to align the real fashion item to the model in the design draft.
Our appearance generation network mainly follows the architecture of $G_d$ introduced in Sec.~\ref{sec:Gr}, containing a saliency-based sampling layer $S$ and a generator $g_a$.}
\firstwwj{The difference is that, we feed both the distortion image $t_a=S(r)$ and the guidance $e$ into $g_a$. For discriminators, $r$ and $e$ are used as the condition.}

To help our network better generate the body part covered by the original clothes,
the most intuitive solution is to rely on the semantic segmentation maps of \firstwwj{the} design draft and treat the body part and clothes part separately. However, existing segmentation \firstwwj{methods have} poor performance on our dataset.
\ys{Instead, we show incorporating high-level labels into our discriminators is able to provide adequate supervision.
Specifically, we design four labels to represent the combination of short or long sleeves for \firstwwj{the} upper body and short or long pants/skirts for \firstwwj{the} lower body.} We label the real fashion items in our dataset. Our main idea is that the output design draft should have the same label $l$. To achieve this, we add an auxiliary classifier on top of $D_a$ to judge the label. $D_a:\{D_a^{tf},D_a^{label}\}$ introduces both the GAN loss and the label loss:
\firstwwj{
\begin{equation}
L_{label}=\mathbb{E}_l(-\text{log} D_a^{label}(l|d))+\mathbb{E}_l(-\text{log} D_a^{label}(l|G_a(r,e))).
\end{equation}
}
Additionally, we propose another discriminator $D_m$ to judge whether we generate the design draft with respect to \firstwwj{the} exemplar $e$. We feed both \firstwwj{the} exemplar $e$ and the result $B_a$ to $D_m$. We choose cross entropy GAN loss for this task.
\firstwwj{The final loss is:}
\begin{equation}
L_a=L_{G}+L_{D_a^{tf}}+L_{D_m}+\lambda_4 L_1
+\lambda_5 L_{per}+\lambda_6 L_{label}.
\end{equation}

\subsection{Fusion Network}
\label{sec:comparison}

As a result of L1 loss and non-pixel-alignment of the body part between $e$ and $d$, the preliminary result $B_a=g_a(t_a,e)$ is kind of blurred.
Luckily we have the structure-aligned distortion image $t_a$.
We want to combine the clothing part from the preliminary result $B_a$ and distorted image $t_a$, and the body part from the exemplar $e$.
We \firstwwj{predict} two masks $m_1$ and $m_2$ through feeding three images into a fusion network $G_f$ \firstwwj{and obtain the final result}:
\begin{equation}
C=B_a\otimes m_1+ e\otimes m_2+t_a\otimes (1-m_1-m_2).
\end{equation}

Similar to our previous fusion network, we apply the contextual loss to capture unaligned appearance and TV loss to make the mask more uniform.
The objective function of our fusion network is
\begin{equation}
L_{f}=L_c+\lambda_6 L_{tv}(m_1)+\lambda_7 L_{tv}(m_2).
\end{equation}

\section{Experimental Results}
\label{sec:ablation}

\subsection{Implementation Details}

\ys{
\textbf{Dataset.}
We collect 1,173 paired $256\times 256$ design drafts and real fashion item images from the display pictures of e-shops.
For design draft to real fashion item translation, 1,023 images are used for training and the \wwj{remaining} 150 \wwj{are} for testing.
For real fashion item to design draft translation, we select seven classes from the dataset, each class contains design drafts \wwj{of} similar \wwj{models} and \wwj{poses}. The biggest class contains 362 images and the smallest class contains 22 images. Seven classes totally contain 717 images for training.
}

\textbf{Network Structure.}
Our detail preservation network, shape correction network and appearance generation network share \wwj{a} similar network structure. The generators of $g_d$, $g_s$ and $g_a$ are \wwj{built} upon U-Net~\cite{isola2017image}. The saliency-based sampling layer follows~\cite{recasens2018learning} and the discriminators follow PatchGAN~\cite{isola2017image}.

\textbf{Training Setup.}
For all the experiments, we set $\lambda_1=50$, $\lambda_2=5$, $\lambda_3=1$, $\lambda_4=50$, $\lambda_5=5$, $\lambda_6=1$, $\lambda_7=1$.
We use minibatch SGD and apply the Adam solver, with a learning rate of 0.0002, and momentum parameters $\beta_1 = 0.5$, $\beta_2 = 0.999$.

\subsection{Comparison with State-of-the-Art Methods}

To validate the effectiveness of the proposed D2RNet and R2DNet, we compare them with the following state-of-the-art image-to-image translation methods:
\ys{
\begin{itemize}
  \item \textbf{Pix2pix} \cite{isola2017image}: The standard image-to-image translation framework, which is the model our network bases on.
  \item \textbf{CycleGAN} \cite{CycleGAN2017}: Because our data is not pixel wisely aligned, we also compare with unsupervised CycleGAN to check the performance if our dataset is simply regarded as unpaired. %
  \item \textbf{Pix2pixHD}  \cite{wang2018pix2pixHD}: A coarse-to-fine image-to-image translation framework for high resolution images.
  \item \textbf{SPADE}  \cite{park2019SPADE}: A latest image-to-image translation framework that leverages spatially-adaptive normalization. We modify its semantic segmentation map inputs to our deign drafts.
  \item \textbf{StarGAN}  \cite{choi2018stargan}: A multi-domain image-to-image translation framework using similar discriminator architecture as our appearance network. Note that its generator uses labels as input, while our model does not require additional labels as input during testing.
  \item \textbf{Pix2pixSC} \cite{pix2pixSC2019}: A latest exemplar-guided image-to-image translation framework. We consider the models in design drafts as consistent style to keep the model unchanged during translation.
\end{itemize}
}

\begin{figure*}[t]
  \centering
  \includegraphics[width=0.88\linewidth]{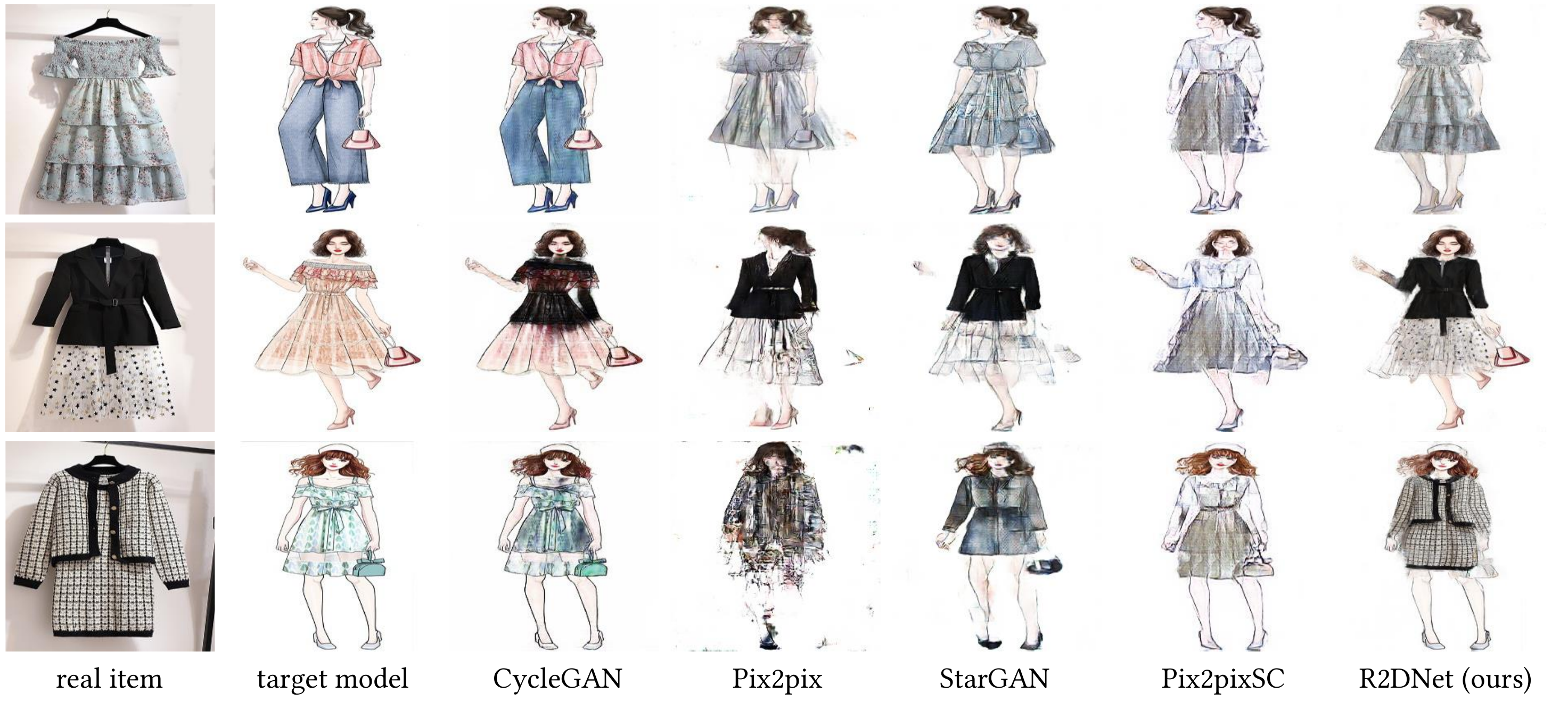}
  \caption{Our R2DNet compared with CycleGAN \cite{CycleGAN2017}, Pix2pix \cite{isola2017image}, StarGAN \cite{choi2018stargan} and Pix2pixSC \cite{pix2pixSC2019}.}
  \label{fig:com2}
\end{figure*}

\begin{figure*}[t]
  \centering
  \includegraphics[width=0.89\linewidth]{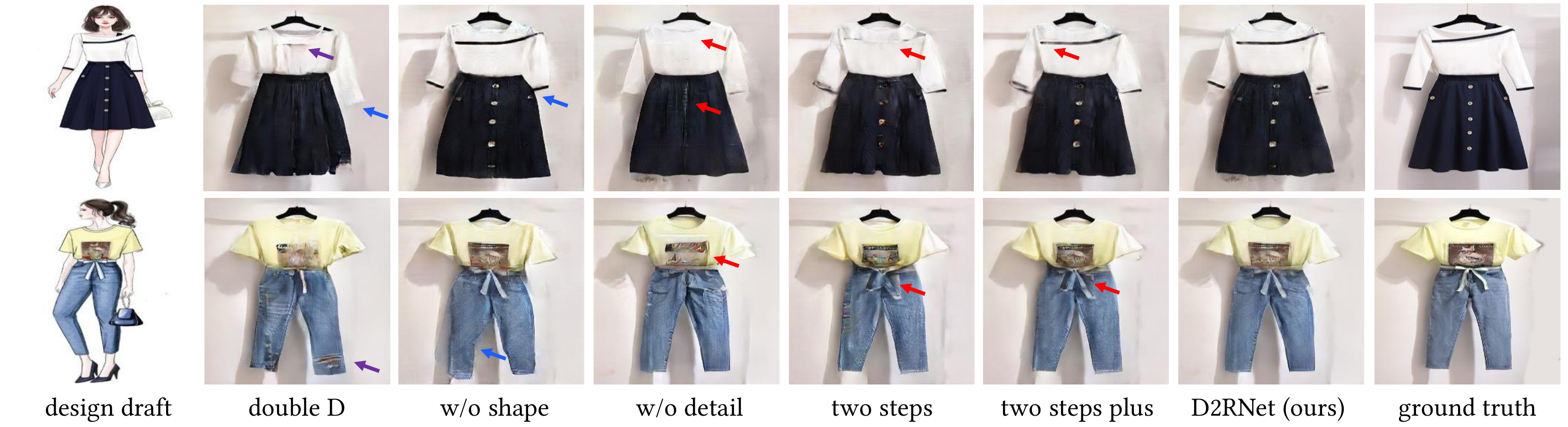}
  \caption{Ablation studies for our D2RNet. The artifacts, the inaccurate shape of the clothes, and missing or blurring textures are indicated by purple, blue and red arrows, respectively.}
  \label{fig:ab1}
\end{figure*}

\begin{figure*}[t]
  \centering
  \includegraphics[width=0.72\linewidth]{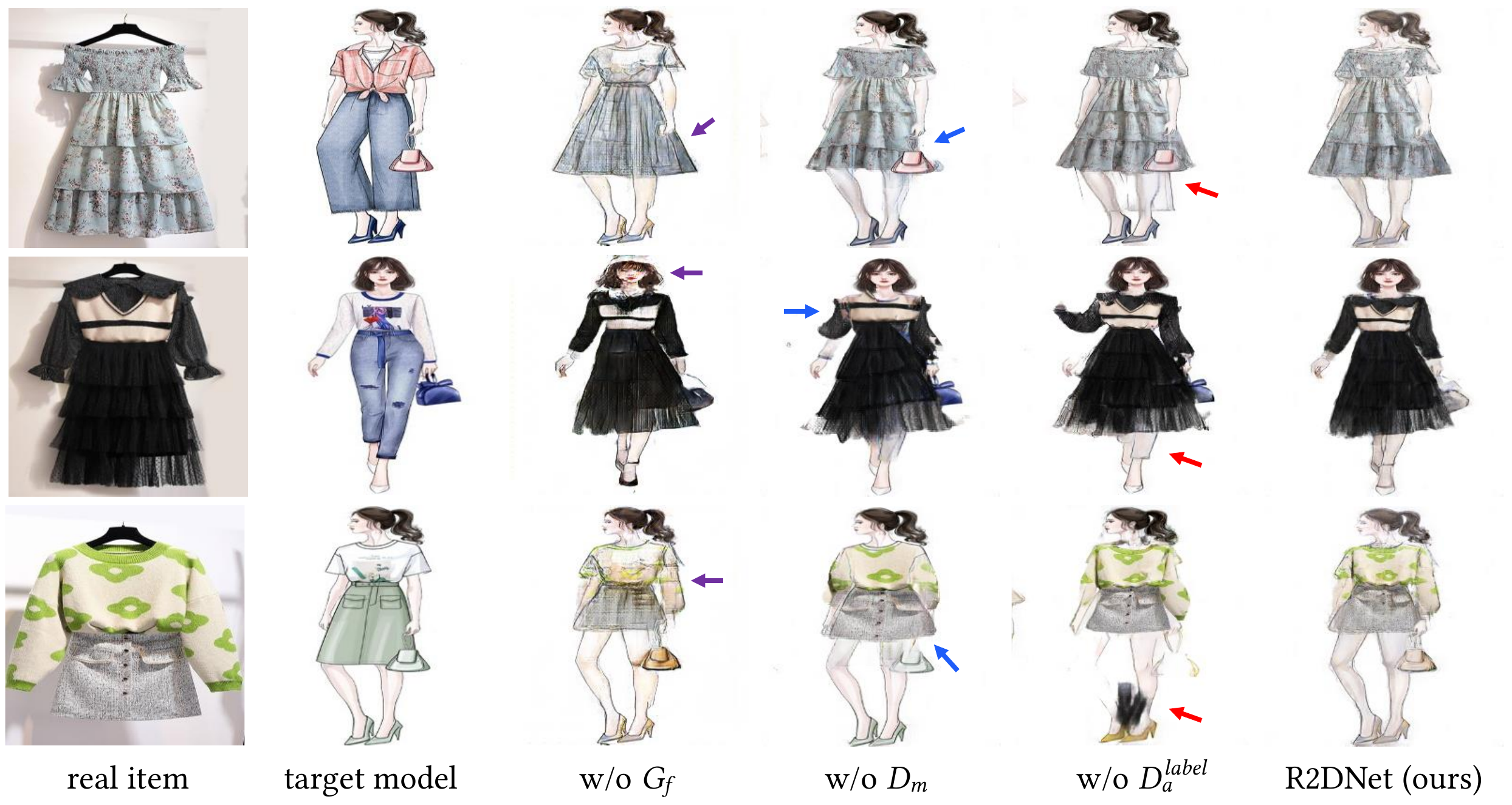}
  \caption{Ablation studies for our R2DNet. The unrealistic textures, the inconsistency between the shape of the clothes and the pose of the models, and the ghosting artifacts are indicated by purple, blue and red arrows, respectively. }
  \label{fig:ab2}
\end{figure*}

\begin{table} [t]
   \caption{Quantitative results on design draft to real fashion item translation. The best user score is marked in bold.}
   \label{tb:d2r}
   \centering
   \resizebox{\linewidth}{!}{
   \begin{tabular}{lccccc}
   \toprule
  Method & CycleGAN & Pix2pix &  Pix2pixHD & SPADE & D2RNet\\
   \midrule
   Scores & 2.07 & 3.29 & 3.50 & 1.20 & \textbf{4.94}\\
   SSIM & 0.459 & 0.631 & 0.608 & 0.548 & \textbf{0.645}\\
   \bottomrule
   \end{tabular}}
   \vspace{0mm}
\end{table}

\ys{\textbf{Qualitative Evaluation}. Fig.~\ref{fig:com1} presents design draft to real fashion item translation results.
For unsupervised method, CycleGAN provides weak constraints for each domain, thus failing to establish accurate mappings between the two domains. For supervised methods, Pix2pix, Pix2pixHD and SPADE heavily rely on the pixel-wise correspondences between the two domains, and can only generate the rough outline of clothes in our unaligned task. By comparison, the proposed D2RNet is superior in generating clear shape of clothes and preserving detailed patterns in design drafts. %
}

\ys{Fig.~\ref{fig:com2} presents real fashion item to design draft translation results. CycleGAN can generate design drafts with only color transferred, which highly resemble the original exemplary target model images. Meanwhile, Pix2pix fails to synthesize the models in the exemplar image and StarGAN does not fit the clothes to the pose of the model, yielding ghosting artifacts. Pix2pixSC fails to characterize the style of clothes, and renders very similar clothes regardless of the input real items. By comparison, the proposed R2DNet can accurately render the clothes on the target models, producing the most satisfying design drafts.}

\ys{\textbf{Quantitative Evaluation}. To quantitatively evaluate our approach, we conducted a user study where users were asked to score the results by five methods on each task. For each task, ten groups of results are shown (for design draft to real fashion item translation task, the ground truth real item images are also shown for reference) and users are tasked to assign 1 to 5 scores to five results in each group based on the visual quality.  A total of 22 users \wwj{participated} and 2,200 scores \wwj{were} collected.  Table \ref{tb:d2r} and Table \ref{tb:r2d} present the mean opinion scores, where the proposed method obtains the highest scores, quantitatively verifying its superiority.} Also, we provide SSIM in Table \ref{tb:d2r} because only this task has ground truth.

  \begin{table} [t]
   \caption{Quantitative results on real fashion item to design draft translation. The best user score is marked in bold.}
   \label{tb:r2d}
   \centering
   \resizebox{\linewidth}{!}{
   \begin{tabular}{lccccc}
   \toprule
  Method & CycleGAN & Pix2pix & StarGAN  & Pix2pixSC & R2DNet\\
   \midrule
   Scores & 1.39 & 3.35 & 3.51 & 2.06 & \textbf{4.69}\\
   \bottomrule
   \end{tabular}}
   \vspace{0mm}
   \end{table}

\begin{figure*}[t]
  \centering
  \includegraphics[width=0.93\linewidth]{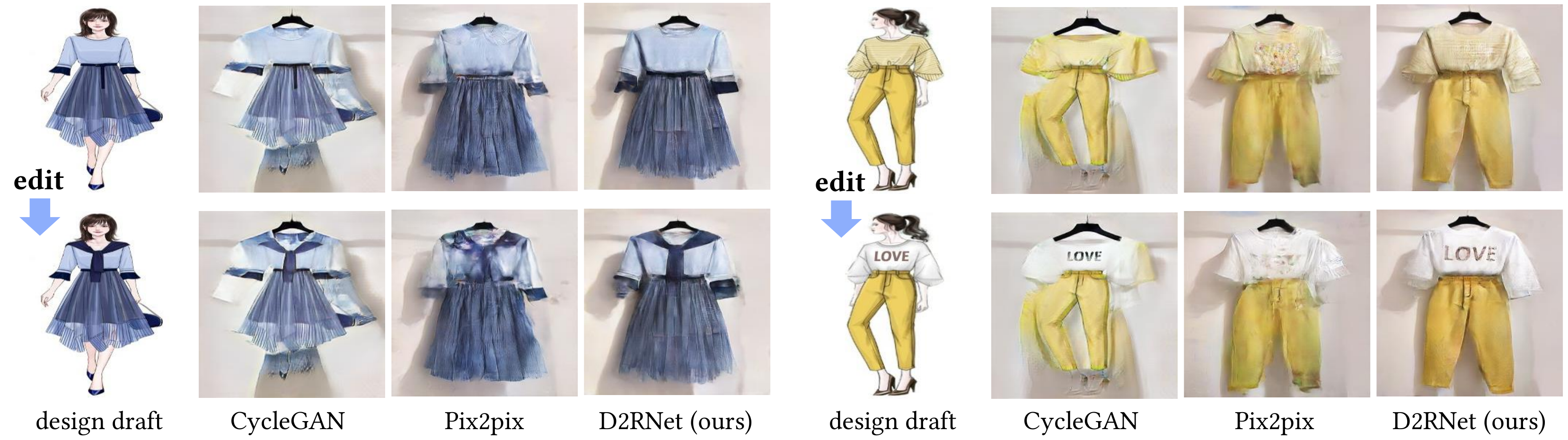}
  \caption{Application in fashion design editing. Top row: input design draft and rendered fashion items by different methods. Bottom row: edited design draft and the corresponding modified fashion items. }
  \label{fig:application}
\end{figure*}

\subsection{Ablation Study}

To analyze each component of D2RNet,  in Fig.~\ref{fig:ab1}, we \wwj{present} the translation results with the following different configurations:

\begin{itemize}
  \item \textbf{Double D}: A saliency-based sampling layer and a U-Net as the generator, which is trained with both $D_d$ and $D_s$.
  \item \textbf{w/o Shape}: Results of our detail preservation network $G_d$.
  \item \textbf{w/o Detail}: Results of our shape generation network $G_s$.
  \item \textbf{Two Steps}: A $G_d$ followed by an additional pix2pix model for shape refinement. The refinement network uses the distortion image and the output of $G_d$ as input. Meanwhile, its discriminator \wwj{aims} to learn to judge the output of $G_d$ as false. $G_d$ is first trained, and is then fixed to train the subsequent refinement network.
  \item \textbf{Two Steps Plus}: A combination results of the output of two steps and the output of $G_d$ using a fusion network.
\end{itemize}

As can be seen in Fig.~\ref{fig:ab1}, without $G_d$, the texture details are missing or blurring, while without $G_s$, the shape of clothes is not properly adjusted. Our \wwj{two-stream} framework effectively solves these two problems by combining $G_d$ and $G_s$.
It can be also observed that directly using two discriminators does not get both benefits. It might be because our saliency-based sampling layer cannot focus on different \wwj{regions} at one time.
In addition, if we use two-step coarse-to-fine networks, as the whole framework goes deeper, the details in the original draft are inevitably and more severely lost.

To analyze each component of R2DNet,  in Fig.~\ref{fig:ab2}, we presents the translation results with the following different configurations:

\begin{itemize}
  \item \textbf{w/o $G_f$}: Results of our appearance generation network $G_a$.
  \item \textbf{w/o $D_m$}: Results of our R2DNet without $D_m$.
  \item \textbf{w/o $D_a^{label}$}: Results of our R2DNet without $D_a^{label}$.
\end{itemize}

In the reverse task, while $G_a$ can warp the clothes to fit the model, its results are rough with plain textures.
Generally, when $D_m$ is removed, the network cannot fit the clothes to model. For example, in the second row of Fig.~\ref{fig:ab2}, the rendered clothes are too loose.  On the other hand, without $D_a^{label}$, the network fails to infer the covered body in the target model, yielding ghosting artifacts.
By comparison, the proposed full R2DNet can well adjust the shape of clothes to fit the target model, and preserve the texture details in the design drafts, showing superior performance.

\subsection{Application}
\ys{Our network allows application of flexible fashion manipulation.
Through our network, fashion designers can manipulate on the design drafts and get their photorealistic form immediately.
As shown in Fig.~\ref{fig:application}, we manually edit the design draft in the top row and get the new fashion items corresponding to our manipulation in the bottom row through the proposed D2RNet. It can be seen that our network performs better than other methods. The edited fashion items have more details corresponding to the modification.}

\section{Conclusions}
\ys{We raise a new problem of translation between design drafts and real fashion items and collected a new paired but unaligned fashion dataset.
We propose a novel D2RNet that translates design drafts to real fashion items and show good performance in both shape preservation to original design drafts and generation of realistic texture details, and a novel R2DNet to solve its inverse task. One interesting future research is to apply our network to other tasks with great shape deformation such as photo-to-caricature translation, photo-to-emoji translation and so on.}

\bibliographystyle{ACM-Reference-Format}
\balance
\bibliography{sample-bibliography}

\end{document}